\documentclass[singlecolumn,letterpaper,12pt]{article}
\usepackage{graphicx}
\usepackage{bm}
\usepackage{graphicx}
\usepackage{latexsym}
\usepackage{amsmath}
\usepackage{amsthm} % pushQED, popQED
\usepackage{times}

\usepackage{txfonts}
\usepackage[bookmarks=false]{hyperref}
\usepackage{stfloats}
\usepackage{algorithm}
\usepackage{algorithmic}

\begin{document}

\date{}

\title{Evaluation of Machine Learning Techniques for Green Energy Prediction}
\author{
Ankur Sahai \\
University of Mainz, Germany
}
%\author{\IEEEauthorblockN{Ankur Sahai}
%\IEEEauthorblockA{College of Computer Science}
%\IEEEauthorblockA{Northeastern University, Boston, MA 02115}
%Email: asahai@ccs.neu.edu}

\hyphenpenalty=100000
\setlength{\parindent}{10pt}
\setlength{\parskip}{1ex} 
\maketitle
\thispagestyle{empty}
\noindent

%%%%%%%%%%%%%%%%%%%%%%%%%%%%%%%%%%%%%%%%%%%%%%%%%%%%%%%%%%%%%%%%%%%%%%%%%%%%%%%%
\section{Objective}
\label{sec: objective}
%%%%%%%%%%%%%%%%%%%%%%%%%%%%%%%%%%%%%%%%%%%%%%%%%%%%%%%%%%%%%%%%%%%%%%%%%%%%%%%%
We evaluate Machine Learning techniques for Green energy (wind, solar and biomass) prediction based on weather forecasts. Weather is constituted by multiple attributes: temperature, cloud cover, wind speed / direction which are discrete random variables. One of our objectives is to predict the weather based on the previous weather data. Additionally we are interested in finding correlation (dependencies in order to reduce the dimensionality of the data set) between these variables, predicting missing data predict deviations in weather forecasts (for job scheduling within the green control center), finding clusters within the data (constituted by closely related variables e.g. PCA that can be used to remove redundant variables), classification, finding (non-linear using SVMs) regression models, training artificial neural networks based on the historical data so that they can be used for prediction in the future. 
%%%%%%%%%%%%%%%%%%%%%%%%%%%%%%%%%%%%%%%%%%%%%%%%%%%%%%%%%%%%%%%%%%%%%%%%%%%%%%%%
\section{Machine learning techniques}
\label{sec:machine-learning-tech}
%%%%%%%%%%%%%%%%%%%%%%%%%%%%%%%%%%%%%%%%%%%%%%%%%%%%%%%%%%%%%%%%%%%%%%%%%%%%%%%%
We analyze the following machine learning techniques:
\begin{itemize}
\item {Bayesian inference}
\item {Neural Networks}
\item {Support Vector Machines}
\item {Clustering technique: PCA}
\end{itemize}
%%%%%%%%%%%%%%%%%%%%%%%%%%%%%%%%%%%%%%%%%%%%%%%%%%%%%%%%%%%%%%%%%%%%%%%%%%%%%%%%
\subsection{Bayesian learning}
\label{sec:bayesian}
%%%%%%%%%%%%%%%%%%%%%%%%%%%%%%%%%%%%%%%%%%%%%%%%%%%%%%%%%%%%%%%%%%%%%%%%%%%%%%%%
Bayesian learning is based on using previous information to predict the future. It is based on the Bayes� rule
\begin{eqnarray}
p(X / Y) = P(X \wedge Y) / P(Y) \\
= P(Y/X)*P(X)/P(Y) \\
\implies p(X / Y) \propto P(X) \\
\implies p(X / Y) \propto (1/P(Y))
\label{operation1}
\end{eqnarray}

If we consider the set of all the events that are correlated to X (i.e. that X depends upon) say $Y_1, \cdots, Y_n$, then from (1) we get:
\begin{eqnarray}
X = (X \wedge Y_1) \vee \cdots \vee (X \wedge Y_n) \\
\implies p(X) = p(X \wedge Y_1) + \cdots + p(X \wedge Y_n) \\
p(X) = p(X/Y_1)*p(Y_1) + \cdots + p(X/Y_n)*p(Y_n) \\
p(X) = \sum_{i= 1}^{n} p(X / Y_i) * p(Y_i)
\end{eqnarray}

Bayesian networks use a directed acyclic graph for the different random variables (represented using nodes) and their conditional dependencies (represented using edges) and uses inference algorithm to make prediction. Different graphical techniques like belief propagation or expectation maximization can be used to spread the information within the Bayesian network. Expectation maximization algorithms are used to model scenarios, which consist of latent variables, and consist of two steps. The structure of a dynamic Bayesian network changes with time.

One simple example is a Hidden Markov Model. Markov model is a graph where each node corresponds to a state and has an associated transition probability to the neighboring nodes (which measure the probability with which the state transition will happen to the neighboring nodes). In the Hidden Markov Model some states are hidden as in the transition probability for these nodes (how the system behaves when in the state corresponding to these nodes) is not known.

Belief propagation includes algorithm like sum-product message passing. This technique is based on using the information stored in the neighboring nodes along with conditional probability to pass message within the network. It can also be represented using a Linear programming formulation. If the input variable follows a distribution such as Gaussian, it becomes much easier to use Bayesian inference. One of the disadvantages is that it is computationally hard although it is conceptually simple.

This technique can also be used to recover the missing parameters / variables based on the previous value (prior distribution) and correlation with other parameters.
%%%%%%%%%%%%%%%%%%%%%%%%%%%%%%%%%%%%%%%%%%%%%%%%%%%%%%%%%%%%%%%%%%%%%%%%%%%%%%%%
\subsection{Clustering techniques}
\label{sec:clustering}
%%%%%%%%%%%%%%%%%%%%%%%%%%%%%%%%%%%%%%%%%%%%%%%%%%%%%%%%%%%%%%%%%%%%%%%%%%%%%%%%
These techniques are used to identify closely related data points in any given data set.  K-means clustering is a commonly studied example where the data points have to be grouped into K clusters such that the mean distance between the clusters is minimized. These techniques can be used to identify patterns in a weather plot say in a temperature vs. cloud cover plot where, region with higher density of points can be identified (which can be used to infer when the temperature is of certain value then it is more likely that the cloud cover is of certain value).

Principal Component Analysis (PCA) is one such technique that is based on generating linearly uncorrelated variable (principal components) from a data set of multiple correlated variables.
%%%%%%%%%%%%%%%%%%%%%%%%%%%%%%%%%%%%%%%%%%%%%%%%%%%%%%%%%%%%%%%%%%%%%%%%%%%%%%%%
\subsection{Binary Decision Trees}
\label{sec:bdd}
%%%%%%%%%%%%%%%%%%%%%%%%%%%%%%%%%%%%%%%%%%%%%%%%%%%%%%%%%%%%%%%%%%%%%%%%%%%%%%%%
This can be used for scheduling the jobs based on deviation between the predicted and actual weather conditions. One can calculate different schedules based upon the different weather forecast / prediction and store it in a binary decision tree. Later on depending on the deviation from the actual weather data one can use the rules stored in the nodes of the decision tree to select one of the schedules.
%%%%%%%%%%%%%%%%%%%%%%%%%%%%%%%%%%%%%%%%%%%%%%%%%%%%%%%%%%%%%%%%%%%%%%%%%%%%%%%%
\subsection{Support Vector Machines}
\label{sec: SVM}
%%%%%%%%%%%%%%%%%%%%%%%%%%%%%%%%%%%%%%%%%%%%%%%%%%%%%%%%%%%%%%%%%%%%%%%%%%%%%%%%
SVM is a supervised machine learning technique, which is used for binary classification of the input into different sets after training the model with previous data. It can work in multiple dimensions represented by different variables. SVM works by building a model which is used for classification.

It is based on a kernel method which is used for classification / regression, reducing dimensionality or clustering. Further they can also be used for non-linearizing the regression and avoiding over / underfitting. This is important because it is not possible to find a linear fit for many of the observed data in real world. In a simple linear regression techniques such as minimizing least square error the decision to select a regression function for the input data is based on fitting a line that minimizes the total square error for the entire data-set. However, for a data set that is very unevenly distributed the linear regression may be not be the best suited technique. Kernel techniques also make it easier to apply linear methods in higher dimensions. Another advantage of Kernel methods is that they allow comparison of two objects consisting of multiple attributes. 

Another perspective to understanding SVMs is to view it as a Minimization (optimization) problem where the goal is to minimize the average distance between the two clusters and the hyperplane which is equivalent to identifying a hyperplane which has the maximum distance from the nearest data point from both the clusters.
LibSVM and SVMLight are two software libraries that can be used for SVM.
%%%%%%%%%%%%%%%%%%%%%%%%%%%%%%%%%%%%%%%%%%%%%%%%%%%%%%%%%%%%%%%%%%%%%%%%%%%%%%%%
\subsection{Artificial Neural Networks}
\label{sec: ANN}
%%%%%%%%%%%%%%%%%%%%%%%%%%%%%%%%%%%%%%%%%%%%%%%%%%%%%%%%%%%%%%%%%%%%%%%%%%%%%%%%
This technique is based on the simulation of a biological neural network, which consists of artificial neurons. By training the artificial neurons over a period of time (with historical data) one can predict future output. It is based on studying how the neurons connect (represented by connection weights) and pass information to each other for different inputs. Neurons collect the input from neighboring neurons and based on the magnitude of this input decide whether to fire or not. There can be multiple layers in a ANN e.g.  input, hidden and output layer.

Additionally we suggest the following techniques: Gibbs sampling, Ridge regression, Mean field value, RBF kernel.

%%%%%%%%%%%%%%%%%%%%%%%%%%%%%%%%%%%%%%%%%%%%%%%%%%%%%%%%%%%%%%%%%%%%%%%%%%%%%%%%
\section{Conclusion}
\label{sec:conclusion}
%%%%%%%%%%%%%%%%%%%%%%%%%%%%%%%%%%%%%%%%%%%%%%%%%%%%%%%%%%%%%%%%%%%%%%%%%%%%%%%%
Bayesian inference can be used to predict the values of missing variables. Clustering techniques can be used to identify similarities between points in a data set in addition to removing redundancy. Binary decision tree can be used for dynamic scheduling (based on rules stored on nodes) within the green control center. SVMs and ANNs have to be analyzed further on if they can be useful for weather prediction.

\end{document}